# rmlnomogram: An R package to construct an explainable nomogram for any machine learning algorithms

*Herdiantri Sufriyana,[1] Emily Chia-Yu Su.[1,2,3,*]*

[1] *Institute of Biomedical Informatics, College of Medicine, National Yang Ming Chiao Tung University, Taipei, Taiwan.*
[2] *Graduate Institute of Biomedical Informatics, College of Medical Science and Technology, Taipei Medical University, Taipei, Taiwan.*
[3] *Clinical Big Data Research Center, Taipei Medical University Hospital, Taipei, Taiwan.*

[*] *Corresponding author:*
*Emily Chia-Yu Su. Institute of Biomedical Informatics, College of Medicine, National Yang Ming Chiao Tung University, 155, Section 2, Linong Street, Beitou District, Taipei 112304, Taiwan. Tel: +886-2-28267000 Ext. 67391. E-mail: emilysu@nycu.edu.tw.*

# Abstract

**Background:** Current nomogram can only be created for regression algorithm. Providing nomogram for any machine learning (ML) algorithms may accelerate model deployment in clinical settings or improve model availability. We developed an R package and web application to construct nomogram with model explainability of any ML algorithms. **Methods:** We formulated a function to transform an ML prediction model into a nomogram, requiring datasets with: (1) all possible combinations of predictor values; (2) the corresponding outputs of the model; and (3) the corresponding explainability values for each predictor (optional). Web application was also created. **Results:** Our R package could create 5 types of nomograms for categorical predictors and binary outcome without probability (1), categorical predictors and binary outcome with probability (2) or continuous outcome (3), and categorical with single numerical predictors and binary outcome with probability (4) or continuous outcome (5). Respectively, the first and remaining types optimally allowed maximum 15 and 5 predictors with maximum 3,200 combinations. Web application is provided with such limits. The explainability values were possible for types 2 to 5. **Conclusions:** Our R package and web application could construct nomogram with model explainability of any ML algorithms using a fair number of predictors.

# Introduction

Application of artificial intelligence is emerging in medicine [1-3]. Particularly, it recently applies more computational machine learning (ML), e.g., random forest, artificial neural network, in addition to statistical ML such as linear or logistic regression for predicting patient outcomes [4, 5]. This phenomenon is driven by larger and more diverse data due to vast digitalization worldwide [6]. Big data take cost, leading more expectation to increase data utilization. It includes either diagnostic or prognostic prediction of patient outcomes, particularly using high-dimensional predictors in electronic health records (EHRs) [7]. Using a sufficient sample size, a prediction model is likely more accurate if it uses high- over low-dimensional predictors. Both are respectively what computational and statistical ML are expected to be good at, as implied by the best practices in their applications [8]. Regardless the number of predictors, there is a tendency to develop a prediction model by selecting the best from both MLs instead of statistical ML alone, as mostly practiced in medicine earlier decades before [1, 4, 9, 10]. The selection may end up with a prediction model applying computational ML. For examples, prediction model studies in medicine were more often selecting the best model by either regression or non-regression algorithms, in which none of them was consistently better than another [11-13]. Yet, a computational ML results in a complex model, except a classification or regression tree (CART) algorithm, which requires integration to EHR system [14].

Complexity probably causes many computational prediction models not yet being deployed in clinical settings [15], or lacking of model availability for public access [16]. In the other hand, a statistical prediction model is mostly, if not all, simpler than a computational one, because statistical ML results in a risk prediction rule, formula, or nomogram. It allows a clinician or another researcher to immediately use or test a prediction model published in a scientific paper, as well as a simple computational ML model such as CART [17]. One may argue a web or mobile application also allows immediate use or test for a computational prediction. Yet, its accessibility is less than a tool that only requires a paper, instead of a computer, a mobile phone, or an internet connection. [18]. It is not seldom that an informatics tool is inaccessible when it is needed, not to mention a costly maintenance to keep a web application running [19]. In this situation, applying a complex prediction model needs a nomogram. However, it is currently only designable for a linear/logistic regression model but not for other ML models.

To solve this problem, we designed a nomogram for any ML prediction models. Particularly, our design currently enables nomogram for a model using minimum a categorical predictor without or with maximum a numerical predictor. While such model does not cover all prediction models, this invention may open a new frontier in ML prediction model deployment studies, i.e., future development of nomogram for other kinds of prediction models. A human user may be also limited in a capability in reading a nomogram with a large number of predictors within a reasonable time. Nevertheless, it is not seldom for a complex ML model using a fairly high dimension but still manually usable. Furthermore, our nomogram design also facilitated model explainability, e.g., the Shapley additive explanation. We developed an R package and web application to construct nomogram with model explainability of any ML algorithms that use categorical without or with single numerical predictors to predict either a binary or continuous outcome.

## Design and development

**Problem formulation**

We wanted to transform an ML prediction model $f_M$ into a nomogram $N$. Let $X = \{x_1, x_2, \cdots, x_k\}$ be the set of predictors for the model $f_M$, where $k$ is the total number of predictors and $X = X_{cat} \cup X_{num}$ such that $|X_{cat}| \geq 1$ and $|X_{num}| \leq 1$. We defined $\Omega$ as the set containing all possible combinations of predictor values in $X$ (Equation 1). Here, $levels(x_i)$ refers to the set of all categories of the $i$-th categorical predictor. Meanwhile, $values(x_j)$ refers to $x_i \in \{min(x_i), min(x_i) + \Delta x, min(x_i) + 2\Delta x, \cdots, max(x_i)\}$ in training set to develop $f_M$, where $\Delta x$ is the rounding precision which depends on the practical application of the nomogram.

For each combination in $\Omega$, the model $f_M$ computes an output $\bar{y}$ which can be denoted as $\bar{y}_\omega$ for each $\omega \in \Omega$. A nomogram creation function $f_N$ takes $\Omega$ and the corresponding computed outputs $\bar{Y} = \langle \bar{y}_\omega : \omega \in \Omega \rangle$ to construct a visual representation as the nomogram $N$ (Equation 2). Optionally, we wanted to incorporate the explainability values for each predictor, e.g., SHAP values, into the nomogram $N$. For each combination in $\Omega$, the explainer of the model $f_M$ computes an explainability value $\phi$ which can be denoted as $\phi_\omega$ for each $\omega \in \Omega$. A nomogram creation function $f_N$ also takes the corresponding computed SHAP values $\Phi = \langle \phi_\omega : \omega \in \Omega \rangle$ to construct a visual representation as the nomogram $N$ (Equation 3).

$$\Omega = \prod_{x_i \in X_{cat}} levels(x_i) \times \prod_{x_j \in X_{num}} values(x_j) \quad \text{Equation 1}$$

$$f_N: \Omega \times \bar{Y} \to N \quad \text{Equation 2}$$

$$f_N: \Omega \times \overline{Y} \times \Phi \to N \quad \text{Equation 3}$$

Note, it is important to reduce the number of predictors before developing the model. Otherwise, the computation will be exponentially more expensive along with the number of predictors. Even if the computation is possible, a user may not be able to use the nomogram due to large number of predictors.

**Nomogram creation algorithm**

This algorithm has been previously described for categorical predictors and binary outcome without probability [20]. The purpose is to reduce the number of combinations of categorical predictors by merging the combinations that belong to the same prediction. First, compute a maximum explainability value for each predictor (step 1 Algorithm 1). For positive prediction, the procedures below are started from predictor with the highest to lowest maximum explainability values (step 3 Algorithm 1). For negative prediction, the procedures below are started from predictor with the lowest to highest maximum explainability values (step 11 Algorithm 1). The procedures to obtain the design are as follow. For the first iteration, select predictors accumulatively from the first to current predictor. For next iterations, filter out samples with the same predictor values with those of previous iteration. For each combination of available predictor values, filter out a combination with minimum and maximum probabilities ≥ and < threshold for positive and negative prediction, respectively (steps 9 and 17 Algorithm 1).

**Algorithm 1. Creation of a nomogram for categorical predictors and binary outcome without probability.**

---

**Require:** $\Omega, \overline{Y}, \Phi, \tau$

1:   $\Phi_{max}(x_i) = \max\limits_{\omega \in \Omega}(\omega, x_i)$
2:   $S = [1, 2, 3, 4, \cdots, k]$
3:   $\Omega \leftarrow sort\_column(desc(\Phi_{max}))$
4:   **for** $s \in S$ **do**
5:      **if** $s = 1$ **then**
6:         $U[s] \leftarrow \Omega[:, 1:s]$
7:      **else**
8:         $U[s] \leftarrow \langle \omega \in \Omega \,|\, \omega_{1:s} \neq U_{-s} \rangle$
9:         $U[s] \leftarrow \langle \omega \in U[s] \,|\, min(\overline{Y}_\omega) \geq \tau \rangle$
10:   $U \leftarrow U[1] \oplus U[2] \oplus \cdots \oplus U[k]$
11:   $\Omega \leftarrow sort\_column(asc(\Phi_{max}))$
12:   **for** $s \in S$ **do**
13:      **if** $s = 1$ **then**
14:         $L[s] \leftarrow \Omega[:, 1:s]$
15:      **else**
16:         $L[s] \leftarrow \langle \omega \in \Omega \,|\, \omega_{1:s} \neq L_{-s} \rangle$
17:         $L[s] \leftarrow \langle \omega \in L[s] \,|\, max(\overline{Y}_\omega) < \tau \rangle$
18:   $L \leftarrow L[1] \oplus L[2] \oplus \cdots \oplus L[k]$
19:   $N \leftarrow tile\_plot(U \oplus L)$

---

If a user does not provide the explainability values $\Phi$, an alternative value is computed for each predictor as a substitute for the maximum explainability value. A univariate regression analysis is conducted for each pair of predictor and output. Subsequently, the upper bound of the odds ratio or estimate is utilized as the alternative value.

**Axis design**

*Categorical predictors and binary outcome without probability*
The filtration results $U$ and $L$ are merged for positive and negative predictions (step 18 Algorithm 1). A tile plot is created in two dimensions (see Figure 1A in "Application examples"). The x-axis is nomogram-reading iterations from the first to the last iteration (left-to-right positioning). Meanwhile, the y-axis is predictors from the highest to lowest maximum explainability values (top-to-bottom positioning). We plotted predictor values with colors of red and cyan for predictors respectively with negative and positive values. The plot panels are divided into positive (left) and negative (right) predictions.

*Categorical predictors and binary/continuous outcome*
This type of nomogram does not require Algorithm 1. The maximum number of categorical predictors depends on the practical application of the nomogram. There is maximum three panels horizontally in this nomogram for (see Figure 1B in "Application examples"): (1) a tile plot of feature value combinations; (2) a line plot of predicted probabilities; and (3) a line plot of feature explainability values. In the first panel, the x-axis is predictors from the highest to lowest maximum explainability values (left-to-right positioning). Meanwhile, the y-axis is predictors with tree-like positioning such that a nomogram reader can easily identify a predictor value adjacent to the previous one according to x-axis. We plotted predictor values with colors of red and cyan for predictors respectively with negative and positive values. In the second panel, the x-axis is the predicted probabilities, while the y-axis is predictors with the same order of predictors with those in the first panel. For binary outcome, the threshold is also shown as a vertical dotted line. Eventually, in the third panel, the x-axis is the explainability values, while the y-axis is predictors with the same order of predictors with those in the first and second panels. Each predictor line is individually color-coded.

*Categorical with single numerical predictors and binary/continuous outcome*

This type of nomogram also does not require Algorithm 1; hence, the maximum number of categorical predictors also depends on the practical application. Similar to nomogram without single numerical predictor, there is maximum three panels horizontally in this nomogram (see Figure 1C in "Application examples"). The only exception is that the y-axis in the second panel is the estimated values instead of the predicted probabilities. Accordingly, no threshold line is shown.

**Web application**

To facilitate user without programming background in R language, we provided an open-access web application to use this package. However, we need to limit the maximum numbers of either categorical predictors or all possible combinations of predictor values. This limitation depends on our workaround for each type of nomogram.

## Comparison

Several tools are available for creating a nomogram. However, all the tools can only create nomograms for regression model. Most of them only accept input as a model that is developed using a specific platform (Table 1).

Table 1. Comparison of currently available tools for creating a nomogram.

| Feature | Tool | | | | | |
| --- | --- | --- | --- | --- | --- | --- |
|  | rmlnomogram | simpleNomo | hdnom | rms | Stata | SAS |
| Programming language | R | Python | R | R | Stata | SAS |
| Package/library | Yes | Yes | Yes | Yes | No | No |
| Web application | Yes | Yes | No | No | No | No |
| Binary outcome | Yes | Yes | No | No | No | No |
| Continuous outcome | Yes | No | No | No | No | No |
| Machine learning algorithm other than regression | Yes | No | No | No | No | No |

## Using Instruction

**Step 1: Prepare input for nomogram creation**

Current procedures for creating the nomogram requires the model to use >1 categorical predictors without or with 1 numerical predictor to predict binary outcome without or with probability or continuous outcome. Predictors with >2 categories should be binarized before model training. Categorizing a numerical candidate predictor should not be determined by an expert panel consensus or a reference from a large-scale, independent dataset, not the same data with those for developing the model, based on the PROBAST guidelines [21].

After developing the model, create a new dataset which comprises all possible combinations of predictor values. If there is a numerical predictor, limit the possible values by determining the minimum, maximum, and rounding precision. Use the new dataset to obtain the predicted probabilities (for binary outcome) or the estimated values (for continuous outcome) and save the results as a single-column table with "output" as the column name. Optionally, obtain the feature explainability values. In our example, we computed the SHAP values. Ensure the datasets for feature values and explainability values have the column names in the same order. In the former dataset, all columns for categorical predictors must use the data type "factor", while the numerical predictor must use the data type "numeric". Therefore, there are three inputs: (1) sample features, a dataset comprising all possible combinations of predictor values; (2) sample output, a single-column dataset containing the predicted probabilities or the estimated values; and (3) feature explainability, a dataset comprising the corresponding explainability value for a predictor and a sample.

**Step 2: Create nomogram**

*Option 1: Use R package*

Install rmlnomogram from CRANs (line 1 Code snippet 1). Alternatively, install the development version of rmlnomogram from GitHub (lines 2 and 3 Code snippet 1). Load nomogram (line 3 Code snippet 1). Load example data for categorical predictors and binary outcome (lines 5 and 6 Code snippet 1). Create the nomogram for this example without and with probability (lines 7 and 8 Code snippet 1). Optionally, load example data for the SHAP values and create the nomogram with probability and SHAP values (lines 9 and 10 Code snippet 1). For categorical with single numerical predictors and binary outcome, load the second set of example data and create the nomogram, accordingly (lines 11 to 14 Code snippet 1). For continuous outcome, we also provided the third and fourth sets of example data with categorical predictors without (lines 15 to 18 Code snippet 1) or with single numerical predictor (lines 19 to 22 Code snippet 1).

**Code snippet 1. Creation of a nomogram using R package.**

```
1:   install.packages("rmlnomogram")
2:   install.packages("devtools")
```

```
 3:    devtools::install_github("herdiantrisufriyana/rmlnomogram")
 4:    library(rmlnomogram)
 5:    data(nomogram_features)
 6:    data(nomogram_outputs)
 7:    create_nomogram(nomogram_features, nomogram_outputs)
 8:    create_nomogram(nomogram_features, nomogram_outputs, prob=TRUE)
 9:    data(nomogram_shaps)
10:    create_nomogram(nomogram_features, nomogram_outputs, nomogram_shaps, prob=TRUE)
11:    data(nomogram_features2)
12:    data(nomogram_outputs2)
13:    data(nomogram_shaps2)
14:    create_nomogram(nomogram_features2, nomogram_outputs2, nomogram_shaps2, prob=TRUE)
15:    data(nomogram_features3)
16:    data(nomogram_outputs3)
17:    data(nomogram_shaps3)
18:    create_nomogram(nomogram_features3, nomogram_outputs3, nomogram_shaps3, est=TRUE)
19:    data(nomogram_features4)
20:    data(nomogram_outputs4)
21:    data(nomogram_shaps4)
22:    create_nomogram(nomogram_features3, nomogram_outputs3, nomogram_shaps3, est=TRUE)
```

*Option 2: Use web application*

We have provided the web application (https://predme.app/rmlnomogram). A user can save the datasets as comma-separated value (CSV) files. To avoid misidentifying a categorical predictor using number as a numerical one, a user must provide an additional, two-column dataset: (1) feature; and (2) category. Upload these files and fill the provided filled to configure the predicted probabilities or the estimated values. If the datasets fulfilled the limits of the maximum numbers of both categorical predictors and all possible combinations of predictor values, then the web application will show the nomogram. After several workarounds, we limit maximum 15 categorical predictors for binary outcome without probability. Meanwhile, only 5 predictors at maximum are allowed for other types of nomograms, with maximum 3,200 combinations.

**Step 3: Read nomogram**

Reading the nomogram is straightforward for the types with the predicted probability or the estimated values. Without probability, i.e., the nomogram that uses Algorithm 1, it requires 3 iterative steps for inferring the prediction, as follow (Figure 1A). These steps have been previously described [20].

1. **Find a predictor with a positive value from the higher to lower maximum impacts; if none, find an iteration at which all available rectangles are negative and jump to step 2a.** In our example, there were 2 of 4 predictors with positive values, i.e., qsec.1 and cyl.6. We started with a positive predictor, i.e., qsec.1.

2. **Match the colors of available rectangles with the predictor values from the earlier to later iterations.** In our example, we matched the colors at iteration 2.
   a. **If they are matched, then draw a vertical line down to identify which panel the iteration belongs to.** However, none was matched for qsec.1 and cyl.6 at iteration 2 (Figure 1A). In this iteration, the rectangle was only available for qsec.1; thus, we did not need to check the values of other predictors.
   b. **If they are not matched, then draw a horizontal line to the next right positive rectangle and repeat step 2 until a rectangle of the next positive predictor is available downward.** The predictor value of qsec.1 was not matched at iteration 2. Thus, we checked the next right positive rectangle until a rectangle of the next positive predictor is available downward, i.e., qsec.1 at iteration 6.
3. **Repeat step 1.** In our example, we repeated step 1 with another positive predictor, i.e., cyl.6. Eventually, this example was matched at iteration 6 which belonged to negative prediction panel.

## Application examples

Here we provided examples for using the R package. We need to install and load necessary packages other than rmlnomogram depending on the platform which is used to create a machine learning model (lines 1 to 4 Code snippet 2). In our example, we develop the model using mtcars (line 5 Code snippet 2), which is widely available, including in base R. This dataset comprises fuel consumption and 10 aspects of automobile design and performance from the 1974 Motor Trend US magazine[22]. We used tidyverse package to help prepare the datasets. Meanwhile, model training in our example applied random forest in caret package and the SHAP values were computed using iml package.

**Binary outcome or class-wise multinomial outcome**

*Example 1: Categorical predictors and binary outcome without probability*
For the first and second examples, we selected transmission (am) for binary outcome and 2 categorical predictors (number of cylinders [cyl] and engine [vs]), and another categorical predictor by categorizing ¼-mile time or qsec (lines 6 to 10 Code snippet 2). The categorical predictors were binarized and the reference categories were removed (lines 10 to 12 Code snippet 2). We trained a random forest model, configuring probability as the output, and created a new dataset comprising all possible combinations of the selected predictors' values (lines 13 to 18 Code snippet 2). The new dataset was used to compute the probability of manual transmission as positive prediction (line 19 Code snippet 2). We created a nomogram for categorical predictors and binary outcome without probability (line 7 Code snippet 1; Figure 1A).

**Code snippet 2. Preparing inputs with categorical predictors and binary outcome.**

```
1:   library(tidyverse)
2:   library(caret)
3:   library(iml)
4:   library(rmlnomogram)
5:   data("mtcars")
6:   mtcars$am <- factor(mtcars$am, levels=c(0, 1), labels=c("Auto", "Manual"))
7:   mtcars$cyl <- factor(mtcars$cyl)
8:   mtcars$vs <- factor(mtcars$vs)
9:   mtcars$qsec <- factor(as.integer(mtcars$qsec >= 18))
10:  mtcars <- select(mtcars, where(is.factor))
11:  dummy_vars <- dummyVars(am ~ ., data=mtcars) |> predict(newdata=mtcars) |> as.data.frame()
12:  mtcars_binarized <- mutate_all(dummy_vars, as.factor) |> mutate(am=mtcars$am) |> select(-vs.0, -cyl.4, -qsec.0)
13:  set.seed(1)
14:  train_config <- trainControl(method="none", classProbs=TRUE)
15:  model <- train(am ~ ., data = mtcars_binarized, method = "rf", trControl=train_config)
16:  caret_features <- str_remove_all(model$finalModel$xNames, "1$")
17:  mtcars_selected <- mtcars_binarized[, caret_features]
18:  nomogram_features <- expand.grid(lapply(mtcars_selected, unique))
19:  nomogram_outputs <- predict(model, nomogram_features, type="prob") |> select(output=levels(mtcars_binarized$am)[2])
20:  X <- mtcars_binarized[, -which(names(mtcars_binarized) == "am")]
21:  predictor <- Predictor$new(model=model, data=X)
22:  nomogram_shaps <- list()
23:  for(i in seq(nrow(nomogram_features))){
24:    shapley <- Shapley$new(predictor, x.interest=nomogram_features[i, ])
25:    nomogram_shaps[[i]] <- shapley$results
26:  }
27:  names(nomogram_shaps) <- seq(nrow(nomogram_features))
28:  nomogram_shaps <- reduce(imap(nomogram_shaps, ~ mutate(.x, i=.y)), rbind)
29:  nomogram_shaps <- nomogram_shaps |> filter(class == levels(mtcars_binarized$am)[2])
30:  nomogram_shaps <- nomogram_shaps |> select(i, feature, phi) |> spread(feature, phi)
31:  nomogram_shaps <- nomogram_shaps |> arrange(as.numeric(i)) |> column_to_rownames(var="i")
```

*Example 2: Categorical predictors and binary outcome with probability*

This example used the same input with that of example 1. If the probability is shown in the nomogram, the explainability values can also be shown as well. We created the explainer using the model and the predictor values of the training set (lines 20 and 21 Code snippet 2). The SHAP values of all samples were computed for each predictor using the explainer (lines 22 to 27). Subsequently, we created the dataset of feature explainability with the same column names and order with those of the dataset of sample features (lines 28 to 31 Code snippet 2). However, the feature explainability consisted the SHAP values for positive prediction, instead of the predictor values. Eventually, we created the nomogram with probability and explainability values (line 10 Code snippet 1; Figure 1B).

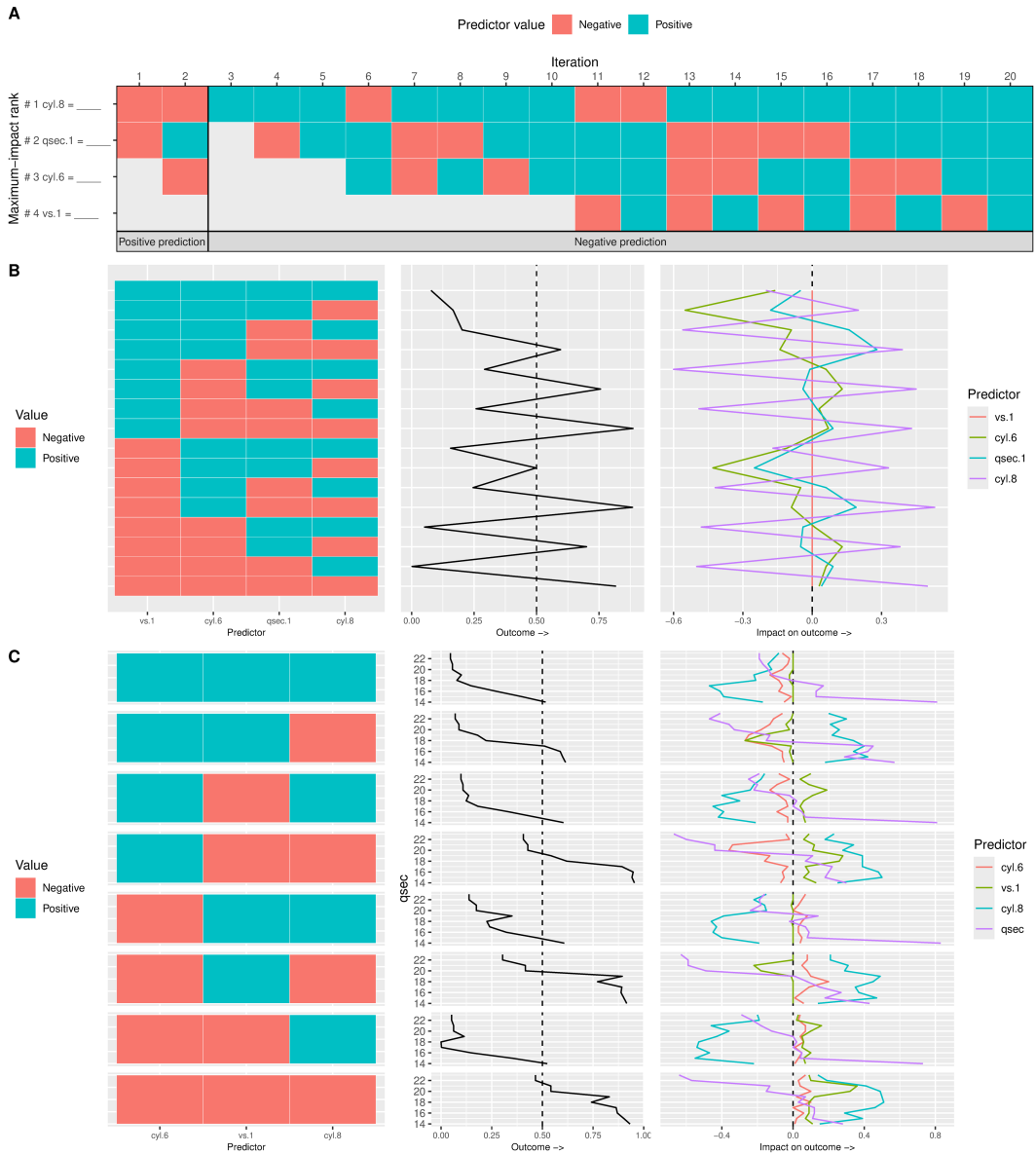

**Figure 1. Nomogram for binary outcome using mtcars dataset: (A) categorical predictors without probability; (B) categorical predictors with probability; and (C) categorical with single numerical predictors with probability.**

*Example 3: Categorical with single numerical predictors and binary outcome with probability*

In this example, we did not categorize qsec, but we set the rounding precision to have no decimal (lines 1 to 2 Code snippet 3). The categorized version of qsec was removed and the numerical one was added (line 3 Code snippet 3). After training the model the same way with examples 1 and 2, we created the dataset of sample features, where qsec ranged from minimum to maximum values in the training set with increment of 1 (lines 4 to 7 Code snippet 3). Similar to example 2, we also created the nomogram with probability and explainability values (Figure 1C).

**Code snippet 3. Preparing inputs with categorical with single numerical predictors and binary outcome.**

```
1:   data("mtcars")
2:   mtcars <- mutate(mtcars, qsec = round(qsec, 0))
3:   mtcars_mixed <- cbind(mtcars["qsec"], select(mtcars_binarized, -qsec.1))
4:   mtcars_selected <- mtcars_mixed[, caret_features]
5:   nomogram_features_num <- select_if(mtcars_selected, is.numeric) |> lapply(\(x) seq(min(x), max(x)))
6:   nomogram_features_cat <- select_if(mtcars_selected, is.factor) |> lapply(unique)
7:   nomogram_features <- expand.grid(c(nomogram_features_num, nomogram_features_cat))
```

**Continuous outcome**

*Example 4: Categorical predictors and continuous outcome*

This example was similar to example 2 but we used weight (wt) for continuous outcome (lines 1 and 2 Code snippet 4). After creating the dataset of sample features the same way with example 2, we created the dataset of sample output (line 3 Code snippet 4). For the explainability values, we applied the same pipeline with example 2, except, we did not filter the SHAP values since for positive prediction (line 29 Code snippet 2). Using these inputs, the nomogram was created with the estimated values (line 4 Code snippet 4; Figure 2A).

**Code snippet 4. Preparing inputs with continuous outcome.**

```
1:   data("mtcars")
2:   mtcars_binarized <- cbind(mtcars["wt"], select(mtcars_binarized, -am))
3:   nomogram_outputs <- data.frame(output = predict(model, nomogram_features))
4:   create_nomogram(nomogram_features, nomogram_outputs, nomogram_shaps, est=TRUE)
```

*Example 5: Categorical with single numerical predictors and continuous outcome*

Similar to examples 3 and 4, we only set the rounding precision of qsec to have no decimal as one of predictors to predict wt for continuous outcome. The datasets of sample features and output were created the same way with examples 3 and 4, respectively. For the explainability values and the nomogram, we also applied the same pipeline with example 4 (Figure 2B).

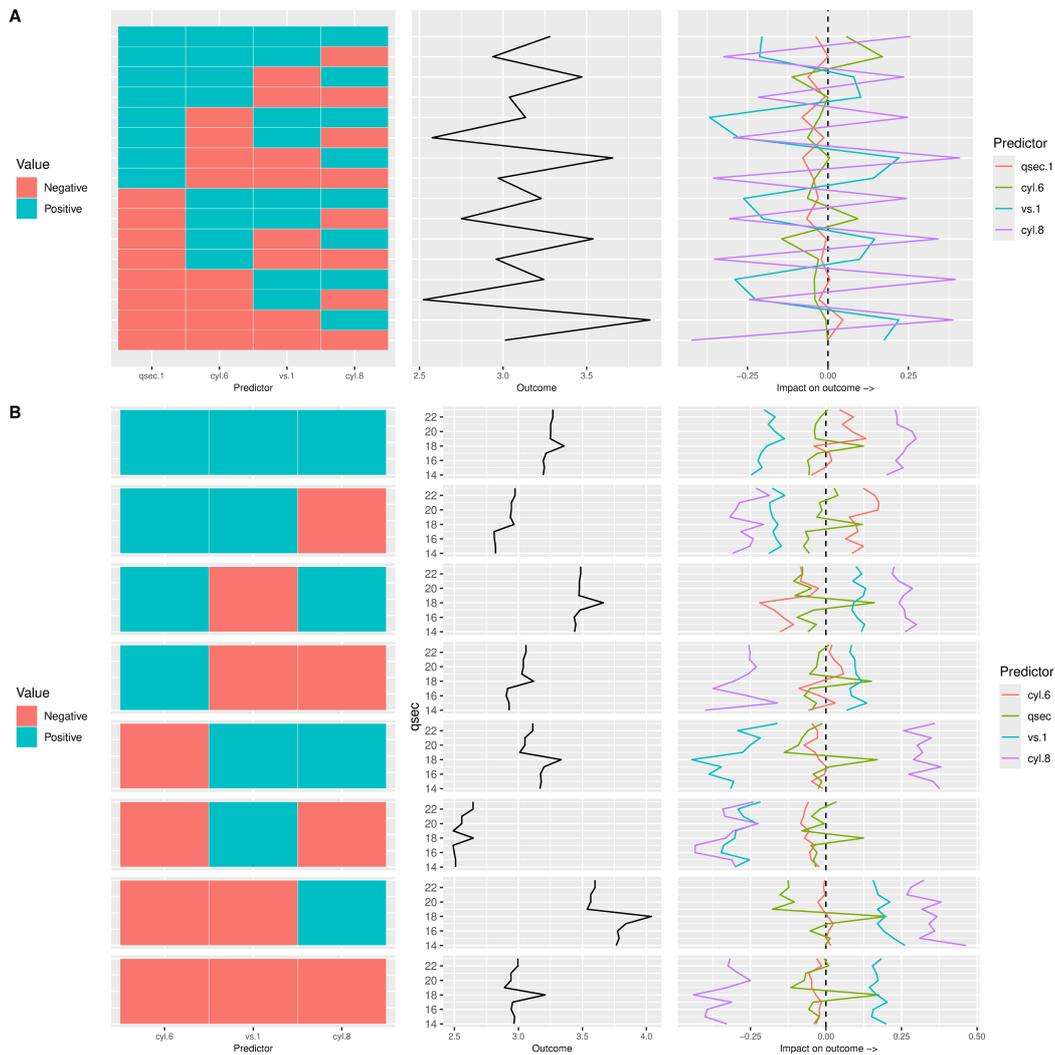

**Figure 2. Nomogram for continuous outcome using mtcars dataset: (A) categorical predictors; and (C) categorical with single numerical predictors.**

# Acknowledgments

This study was funded by: (1) the Postdoctoral Accompanies Research Project from the National Science and Technology Council (NSTC) of Taiwan (grant nos.: NSTC111-2811-E-038-003-MY2 and NSTC113-2811-E-A49A-003) to HS; and (2) the National Science and Technology Council in Taiwan (grant no. NSTC113-2221-E-A49-193-MY3), the Ministry of Science and Technology (MOST) of Taiwan (grant nos.: MOST110-2628-E-038-001 and MOST111-2628-E-038-001-MY2), the University System of Taipei Joint Research Program (grant no.: USTP-NTOU-TMU-112-04), and the Higher Education Sprout Project from the Ministry of Education (MOE) of Taiwan (grant no.: DP2-111-21121-01-A-05 and DP2-TMU-112-A-13) to ECYS. These funding bodies had no role in the study design; in the collection, analysis, and interpretation of data; in the writing of the report; or in the decision to submit the article for publication.

## CRediT authorship

**HS:** Conceptualization, Methodology, Software, Validation, Formal analysis, Writing – original draft, Visualization, Funding acquisition. **ECYS:** Conceptualization, Methodology, Resources, Writing – review & editing, Supervision, Funding acquisition. All authors have read and approved the manuscript and agreed to be accountable for all aspects of the work in ensuring that questions related to the accuracy or integrity of any part of the work are appropriately investigated and resolved.

## Conflicts of interests

The authors declare that they have no competing interests.

## Data availability

We publicly shared the source codes (https://github.com/herdiantrisufriyana/rmlnomogram). An open-access web application (https://predme.app/rmlnomogram) is also provided. Data are provided in the source codes and web application.